\documentclass[letterpaper,10pt,conference]{ieeeconf}
\IEEEoverridecommandlockouts
\usepackage{amsmath,amssymb,amsfonts}
\usepackage{algorithmic}
\usepackage[pdftex]{graphicx}
\usepackage{textcomp}
\usepackage{xcolor}

\usepackage[bookmarks=true]{hyperref}

\makeatletter
\let\NAT@parse\undefined
\makeatother
\usepackage[numbers,sort,compress]{natbib}

\usepackage{siunitx}

\usepackage{xparse}

\newcommand{\obj}[1]{#1}
\newcommand{\pred}[1]{\texttt{#1}}
\NewDocumentCommand \prop {>{\SplitList{ }} m} {\proposition #1}
\NewDocumentCommand \proposition {g g g g} {\texttt{#1}(#2
  \IfValueTF{#3}{,\,#3}{}
  \IfValueTF{#4}{,\,#4}{}
  )
}

\newcommand{\func}[2]{\mathop{}#1\left(#2\right)}

\newcommand{\given}{\;\middle|\;}

\hypersetup{
colorlinks=true,
urlcolor=blue,
}

\usepackage[font=footnotesize,justification=justified,singlelinecheck=false]{caption}
\usepackage{etoolbox}
\makeatother
\def\tablename{Table}

\usepackage[all=normal,paragraphs=tight,floats=tight]{savetrees}
\title{\LARGE \bf
Symbolic State Estimation with Predicates \\
for Contact-Rich Manipulation Tasks
}

\author{Toki Migimatsu$^{1, 2}$, Wenzhao Lian$^{1}$, Jeannette Bohg$^2$, and Stefan Schaal$^{1}$
\thanks{$^{1}$Intrinsic Innovation LLC in CA, USA. {\{wenzhaol, sschaal\}@intrinsic.ai}.}%
\thanks{$^{2}$Stanford Artificial Intelligence Laboratory, Stanford University in CA, USA.  \{takatoki, bohg\}@cs.stanford.edu. This research was conducted during Toki's internship at Intrinsic.}%
}

\begin{document}

\maketitle
\thispagestyle{empty}
\pagestyle{empty}


\begin{abstract}
Manipulation tasks often require a robot to adjust its sensorimotor skills based on the state it finds itself in. Taking peg-in-hole as an example: once the peg is aligned with the hole, the robot should push the peg downwards. While high level execution frameworks such as state machines and behavior trees are commonly used to formalize such decision-making problems, these frameworks require a mechanism to detect the high-level symbolic state. Handcrafting heuristics to identify symbolic states can be brittle, and using data-driven methods can produce noisy predictions, particularly when working with limited datasets, as is common in real-world robotic scenarios. This paper proposes a Bayesian state estimation method to predict symbolic states with predicate classifiers. This method requires little training data and allows fusing noisy observations from multiple sensor modalities. We evaluate our framework on a set of real-world peg-in-hole and connector-socket insertion tasks, demonstrating its ability to classify symbolic states and to generalize to unseen tasks, outperforming baseline methods. We also demonstrate the ability of our method to improve the robustness of manipulation policies on a real robot.
\end{abstract}

\section{Introduction}
\label{sec:intro}
Solving robotic manipulation tasks robustly under perception uncertainty is a challenging problem. State estimation methods seek to solve this problem by predicting the ground truth state of the environment from noisy observations~\cite{thrun2002probabilistic,endres2012evaluation,kerl2013dense,wuthrich2013probabilistic}. A typical state representation is object poses~\cite{chhatpar2003localization,andre2016reliable,wirnshofer2019state,thomas2007multi,nottensteiner2021towards}, which are continuously estimated and then used to guide the robot's motion, e.g., for aligning the end-effector with a door knob. However, less attention has been paid to estimating high-level symbolic states, such as whether the door knob is locked or the door is fully shut. Perception of such symbolic states contributes to the robot's functional understanding of the environment and can help decide when a sub-task has succeeded, whether a failure has occurred, and what action to execute next~\cite{zachares2020interpreting}. As the core building block of high-level execution models, symbolic state representations are widely used in state machines, behavior trees~\cite{colledanchise2018behavior}, task planners~\cite{mcdermott1998pddl}, and Robust Logical-Dynamical Systems~\cite{paxton2019representing}.

A common approach to estimating symbolic states is using hand-tuned thresholds as transition conditions~\cite{johannsmeier2019framework}. For example, a state machine for placing a mug on the table might involve lowering the mug until the robot senses a force exceeding 2N at its end-effector, at which point the robot would open its gripper. However, this approach is brittle and prone to failure. If the detected force exceeds 2N due to sensor noise or unforeseen disturbances while the mug is in free space, the robot might open its gripper and drop the mug.

\begin{figure}
    \centering
    \includegraphics[width=\columnwidth]{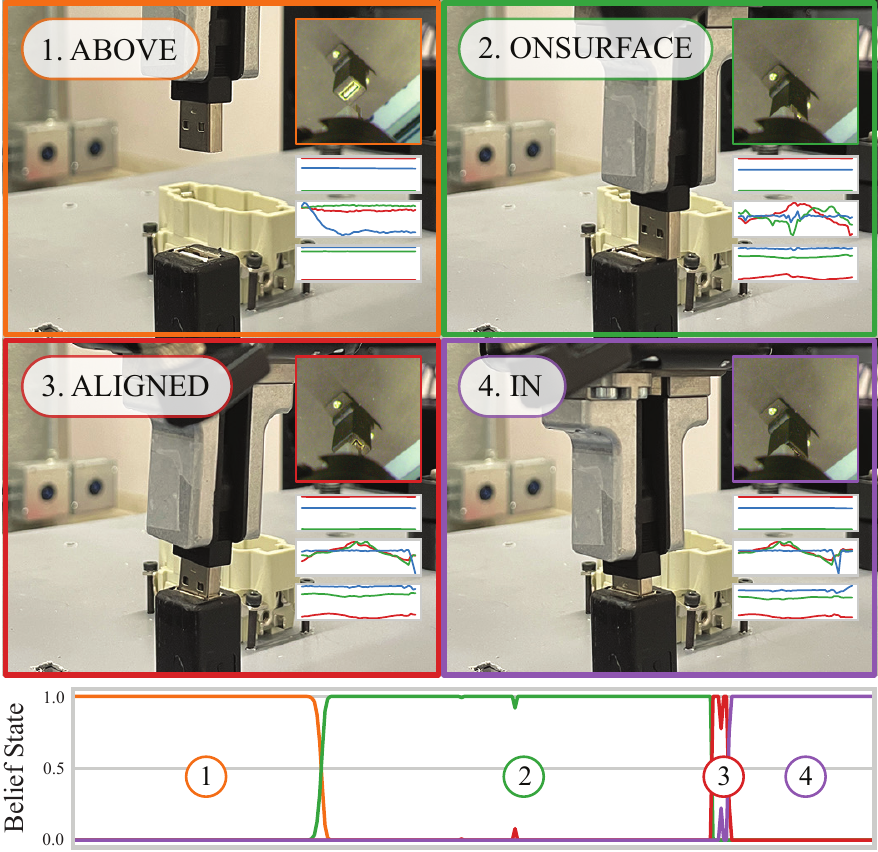}
    \caption{We propose a symbolic state estimation method with predicates for manipulation tasks. Predicate classifiers, trained with a small dataset, detect symbolic attributes from multimodal sensor inputs.
    The noisy classifier outputs are then fed to a Bayesian state estimator to predict symbolic states.
    The execution trajectory illustrated here shows our method applied to a connector insertion task using a predefined set of manipulation primitives. The symbolic state estimator decides when to transition between primitives and identifies failure states. Top: the representative frames in each symbolic state overlaid with the image, position, velocity, and force signals (with RGB lines representing XYZ axes). Bottom: the estimated probability distribution over the symbolic states.}
    \label{fig:teaser}
\end{figure}

\citet{kappler2015data} proposed a data-driven approach for symbolic state estimation. They introduced a Bayesian framework that can combine observations from multiple sensor modalities and detect failures. However, as the state classifiers are trained for each state in the task-specific state machine, this method is hard to scale to unseen tasks and failure cases. \citet{haarnoja2016backprop} proposed using a virtual neural network sensor to encode high-dimensional images to low-dimensional vectors. They train this embedding end-to-end by supervising the posterior estimate of a Kalman filter and differentiating through the recursive update rules. While this differentiable filter has shown promising results with continuous states \cite{kloss2021train}, it has not yet been applied to discrete symbolic states. We will demonstrate that a straightforward adaptation of differentiable filtering to symbolic states underperforms when data is limited and costly to acquire, which is common in real-world robotic manipulation tasks.

In this work, rather than reducing high-dimensional observations to symbolic states directly, we propose reducing to the space of predicates: atomic binary properties such as $\prop{above a b}$ or $\prop{near a b}$ that compose symbolic states. We then learn the noise characteristics of the predicates and use this information to perform Bayesian inference on the symbolic states. There are two main benefits for using predicates as the intermediate representation.
1) As predicates are binary, learning predicate classifiers is easier than learning multi-class state classifiers, thus requiring less data. Further, these atomic predicate classifiers, if trained on a diverse dataset, are generalizable to unseen tasks in the same task family. 2) Since predicates are modular, they can be composed and shared between tasks; adding or removing states from the state space does not require retraining all the predicate classifiers.

Our core contribution is a data-efficient Bayesian framework to perform symbolic state estimation using predicates, handling high-dimensional, multi-modal observations. We evaluate our framework on the peg-hole and connector-socket insertion tasks from National Institute of Standards and Technology (NIST) Assembly Task Board I \cite{lian2021benchmarking} (see Fig.~\ref{fig:teaser}). We collected and labeled a relatively small amount of observations from $120$ open-loop policy executions across $8$ different tasks, and compared methods in two ways: accuracy of offline symbolic state estimation and success rate of online closed-loop policy execution. Experimental results indicate our method achieves the highest state estimation accuracy and task completion rate, outperforming other baselines.

\section{Related Work}
\label{sec:related}
\subsection{Bayesian State Estimation}

Bayesian state estimation is commonly used to estimate continuous states, including robot and object poses, given noisy observations such as range data~\cite{thrun2002probabilistic,endres2012evaluation,kerl2013dense,wuthrich2013probabilistic}. While performing state estimation with RGB images can be difficult due to their high dimensionality and sensitivity to lighting conditions, recent works have proposed differentiable filtering~\cite{kloss2021train}, which uses neural networks to encode images to lower dimensional vectors. These methods embed the recursive Bayesian estimation algorithms (e.g., Kalman~\cite{haarnoja2016backprop}, histogram~\cite{jonschkowski2016end}, and particle filters~\cite{jonschkowski2018differentiable}) into the neural network architecture, enabling end-to-end learning for visual state estimation. \cite{lee2020multimodal} further built upon differentiable filtering to perform sensor fusion. While differentiable filtering has shown promising results for continuous pose estimation where state transitions are continuous, it has not yet been studied extensively for symbolic state estimation, where state transitions are sparse and discrete. \cite{zachares2020interpreting} applied differentiable histogram filtering to estimate a categorical variable. This variable stays constant throughout an episode, and thousands of simulated demonstration runs were collected to train the filter without finally demonstrating it on real data. We show in our experiments that differentiable filtering is not as effective for symbolic state estimation when the amount of training data available is limited---$120$ runs in our case.

\subsection{State Estimation for Assembly Tasks}

State estimation has been applied to assembly tasks in prior works primarily using force data~\cite{chhatpar2003localization,andre2016reliable,wirnshofer2019state}. \cite{thomas2007multi} and \cite{nottensteiner2021towards} fuse visual observations with force data to track part poses, but both methods rely on manually defined image features such as line or blob detectors. All these methods adopt particle filtering to estimate the hole position. However, a significantly large number of particles is required to handle the nonlinear contact dynamics, making this class of methods computationally expensive, sometimes too slow to run in real time~\cite{nottensteiner2016narrow}. We assume that a search strategy can find the hole with high probability given a rough initial guess of its position, as in \cite{nguyen2019probabilistic}. Then, we aim to estimate high-level states such as ``on-surface" and ``fallen" to determine {\em which} high-level action to execute and {\em when}.

\cite{rojas2016robot,rojas2017online} proposes to leverage both pose- and wrench-based trajectory features to identify failure states for assembly. However, they rely on predetermined time intervals to switch between high-level states during task execution. The most relevant work to ours is \cite{kappler2015data}, which proposes a Bayesian symbolic state estimation framework based on force signals and visual data in the form of tracked object positions. However, \cite{kappler2015data} directly trained symbolic state classifiers for each state in a task-specific state machine. This limits the data reuse across tasks and requires model retraining from scratch for novel tasks, objects, and sensor modalities. In contrast, we introduce predicates as the intermediate representation, which improves data efficiency and allows better generalization.

\subsection{High-Level Failure Recovery}

One benefit of performing state estimation on high-level states is the ability to recover from high-level failures, such as the peg falling off the contact surface. \cite{kappler2015data} proposes to use a Support Vector Machine for failure classification, which runs in parallel with their state estimation module and interrupts with a recovery state machine when failures are identified. \cite{lagrassa2020learning} also proposes a separate failure detection module, but their recovery is handled by a model-free policy trained via expert demonstration.  Our framework, on the other hand, includes failure states in the main state space, which unifies the state and failure classification problems into one integrated pipeline.

\section{Bayesian Symbolic State Estimation}
\label{sec:bayesian}
Symbolic states abstract away geometric information and characterize the functional properties of the environment, such as whether two objects are aligned, whether an object is open or closed, etc. We aim to learn a symbolic state predictor that is robust to sensor noise and can handle high-dimensional observations such as images with a minimal amount of training data, suitable for real world robotics applications. Once learned, the state estimator can be used to close the loop on any high-level execution model, including state machines, as demonstrated in Sec.~\ref{sec:close-loop}. 

Inspired by \cite{haarnoja2016backprop}, we build on Bayesian state estimation with virtual sensors to encode high-dimensional observations to low-dimensional vectors. To achieve better data efficiency, we propose training the virtual sensors to output binary predicates rather than an implicit representation of symbolic states learned end-to-end. The predicates---as used in the Planning Domain Description Language (PDDL)~\cite{mcdermott1998pddl}---are defined as binary atomic properties composing symbolic states. After learning virtual sensors for predicates, we then fit generative observation models, e.g., Gaussian Mixture Models (GMMs), to the sensor outputs. The generative models help smooth out prediction errors from the noisy predicate sensors. Although we choose GMMs in our implementation, we leave optimizing the model choice to future work while focusing on the framework design and process integration. The two-step training process outlined above allows us to obtain robust symbolic state estimators for high-dimensional observations, while requiring only a small robot-environment interaction dataset.

\subsection{Domain Specification}
\label{subsec:domain}

The domain for our Bayesian framework can be described by the 6-tuple $\langle S, \Phi, A, T, \Omega, O \rangle$, where $S$ is the set of symbolic states, $\Phi$ is the set of predicates that compose the symbolic states, $A$ is the set of actions, $T$ is the set of state transition probabilities, $\Omega$ is the set of observations, and $O$ is the set of conditional observation probabilities.

\subsection{Belief States}

A belief state represents the agent's estimate of the symbolic state as a probability distribution over the states. At each time step, we can update the belief state $\func{b}{s}$ with
\begin{align}
    \func{b}{s}
        &\propto \func{O}{o \given s, a} \sum_{s_{prev}} \func{T}{s \given s_{prev}, a} \func{b}{s_{prev}}, \label{eq:belief}
\end{align}
where $s_{prev}, s \in S$ are unobservable symbolic states, $a \in A$ is the action performed at $s_{prev}$ to transition to $s$ with probability $\func{T}{s \given s_{prev}, a}$, and $o \in \Omega$ is the observation received at $s$. We estimate $\func{T}{s \given s_{prev}, a}$ and the prior belief state $b(s_0)$ from state visit counts in the training data. The conditional observation probability $\func{O}{o \given s, a}$ can be difficult to compute with traditional estimation methods without a parameterized distribution, since it requires normalizing over the entire observation space. To circumvent this issue, we use binary predicate classifiers as an intermediate representation, serving as a virtual sensor to reduce the dimensionality of observations.

\subsection{Virtual Predicate Sensors}

Predicates $\phi\in\Phi$ define binary properties of objects, usually summarizing geometric information, such as $\prop{inside a b}$ to indicate that object $\obj{a}$ is inside $\obj{b}$. Symbolic states $s \in S$ are defined as sets of predicates $\{\phi_1^s, \phi_2^s, \cdots, \phi_{K^s}^s\}$, where ${K^s}$ is the number of predicates determined by $s$. For example, the symbolic state representing when a robot manipulator has fully inserted a peg into a hole could be $s_{inserted} \equiv \{ \prop{inside peg hole}, \neg \prop{above peg surface} \}$.

We train virtual predicate sensors as binary classifiers $h_{\phi}(o)$ that each outputs a noisy estimate of the probability that predicate $\phi$ is true. The symbolic state $s$ can then be inferred from noisy ``virtual observations" $[h_{\phi_1^s}(o), h_{\phi_2^s}(o), \cdots, h_{\phi_{K^s}^s}(o)]$.

Symbolic states do not need to specify the truth value of all predicates. Given a ground truth state $s$, we thus train the predicate sensors with a variant of the cross entropy loss:
\begin{align}
    L
        &= \sum_{k = 1}^{K^s} \phi_k^s \log h_{\phi_k^s}(o) + (1 - \phi_k^s) \log \left(1 - h_{\phi_k^s}(o) \right).
\end{align}
Rather than computing the cross entropy for all the predicates, we only compute it for the predicates whose truth values are determined by the ground truth state $s$. In practice, this can be implemented by using a boolean mask $\{0, 1\}^{|\Phi|}$ to mask out predicate predictions for predicates not determined by $s$.

\subsection{Virtual Predicate Observation Models}

If a virtual sensor does not output parameters for a parameterized probability distribution, the observation probability $\func{O}{h_{\phi}(o) \given s, a}$ needs to be normalized over its outputs $h_{\phi}(o)$. \cite{jonschkowski2016end} suggests approximating this normalization step with sampling, which is computationally expensive. If the virtual sensors are trained to output categorical variables, as is the case with our predicate sensors, then one could also approximate the normalization by summing over the categorical variables~\cite{zachares2020interpreting}. Using this approximation, however, discards information contained in the continuous probabilities output by the predicate sensors.

For example, suppose a noisy predicate sensor outputs $h_\phi \sim \mathcal{TN}(0.2, 1)$ when $\phi = false$ and $h_\phi \sim \mathcal{TN}(0.4, 0.3)$ when $\phi = true$, where $\mathcal{TN}$ denotes a normal distribution truncated within $[0, 1]$. If we discretize the sensor outputs into two bins, $(h_\phi < 0.5) \Rightarrow (\phi = false)$ and $(h_\phi \geq 0.5) \Rightarrow (\phi = true)$, then observing $h_\phi = 0.4$ leads to predicting $\phi = false$, since $h_\phi < 0.5$. However, the predicate is more likely to be $true$, since $0.4$ is the mean of $h_\phi$ when $\phi = true$.

To capture the fidelity of continuous predictions without resorting to sampling, we represent the observation model with GMMs fit to the virtual predicate sensor output logits:
\begin{align}
    \func{O}{h_{\phi}(o) \given s, a}
        &= \func{GMM_{s, a}}{\log h_{\phi}(o)}.
\end{align}

Because GMMs are parameterized probability distributions, the normalization over the observation space can be computed in closed form. Fitting the GMMs over the same dataset used to train the predicate classifiers could result in overfitting, so we instead use the validation set.

\subsection{Sensor Fusion}

We follow the conventional approach in Bayesian state estimation to integrate observations from multiple sensors. Each sensor observation is considered conditionally independent given the state and action, and thus the conditional observation probability $\func{O}{o \given s, a}$ is computed as the product of all observation probabilities $\func{O}{o_{sensor} \given s, a}$. In our framework, there is one virtual sensor per predicate, so the set of sensors is equivalent to the set of predicates:
\begin{align}
    \func{O}{o \given s, a}
        &= \prod_{\phi} \func{O}{h_{\phi}(o) \given s, a}. \label{eq:fusion}
\end{align}

 To integrate virtual predicate sensors from multiple physical sensor modalities, we can simply define a predicate for each sensor modality. For example, suppose we want to identify a symbolic state where the robot end-effector is in contact with a surface from both visual and force observations. We can introduce two predicates, $\pred{visual-in-contact}$ and $\pred{force-in-contact}$, and define the contact symbolic state to be the conjunction of these two predicates $\{\pred{visual-in-contact}, \pred{force-in-contact}\}$.

\section{Predicate Classifiers}
\label{sec:classifiers}
 In this section, we describe our choice of predicate classifier architectures. However, our proposed framework is agnostic to the classifier implementation; the only requirement is that the classifier outputs a value between 0 and 1.

\subsection{Motion and Force-based Classifiers}

To minimize the amount of data required to train classifiers for motion and force-based predicates, we use logistic regression on a set of handcrafted features. For example, to detect the predicate $\prop{in-contact a b}$, we can simply use the force magnitude as a feature. 
It is possible to train neural networks to classify the predicates without handcrafted features \cite{zachares2020interpreting}, but the predicates for our real-world task are simple enough that doing so would be unnecessary and perform worse in generalization given our small dataset, as we show in the experiments with the state classification baseline. The classifier performance is shown in the top four rows of Table~\ref{tab:predscores}.

\subsection{Image-based Classifiers}
\label{sec:simclr}

For image-based predicate classification, we adopted SimCLRv2 \cite{chen2020big} with a ResNet-50 model \cite{he2016deep} as the backbone network, and added a linear layer on top for predicate classification. We chose SimCLRv2 as it has been shown to be effective for fine-tuning classifiers by augmenting small datasets with random image transformations. The network is pre-trained on ImageNet \cite{deng2009imagenet}, and the linear layer is fine-tuned on our collected dataset of $47,752$ images collected over $120$ policy execution runs across $8$ tasks. We use random cropping and color distortion for data augmentation. The classifier performance is shown in the bottom five rows of Table~\ref{tab:predscores}.

\begin{table}
    \centering
    \footnotesize
    \begin{tabular}{l|c|c|c}
        Predicate & Acc. & Prec. & Recall \\
        \hline
        $\prop{motion-force-axis-aligned a b}$ & 0.96 & 0.97 & 0.98 \\
        $\prop{motion-force-dropping a}$ & 0.99 & 0.91 & 0.58 \\
        $\prop{motion-force-fully-inserted a b}$ & 0.92 & 0.97 & 0.6 \\
        $\prop{motion-force-in-contact a b}$ & 0.97 & 0.96 & 0.98 \\
        $\prop{visual-above a b}$ & 0.95 & 0.93 & 0.92 \\
        $\prop{visual-below a b}$ & 0.85 & 0.95 & 0.84 \\
        $\prop{visual-fallen a}$ & 0.99 & 0.98 & 0.99 \\
        $\prop{visual-fully-inserted a b}$ & 0.94 & 0.93 & 0.79 \\
        $\prop{visual-inserted a b}$ & 0.84 & 0.63 & 0.93 \\
        \hline
        Overall & 0.95 & 0.93 & 0.92 \\
    \end{tabular}
    \caption{Predicate classifier test scores. The simplicity of the atomic predicates allows the classifiers to achieve high test accuracy even with a small train set.}
    \label{tab:predscores}
\end{table}

\section{Real World Environment}
\label{sec:environment}
\subsection{NIST Insertion Tasks}

To evaluate our Bayesian symbolic state estimation framework, we apply it to the insertion tasks on the NIST Assembly Task Board I~\cite{lian2021benchmarking}. We use a subset of the insertion tasks, illustrated in Fig.~\ref{fig:nist}: four connectors (D-sub, RJ45, USB, and waterproof) and four pegs (8x7 rectangle, 12x8mm rectangle, 12mm round, and 16mm round). The task board is rigidly mounted on a worktable, as shown in Fig.~\ref{fig:teaser}. Our hardware system consists of a Kuka IIWA7 arm, Robotiq Hand-E gripper, ATI Mini45 ForceTorque sensor, and wrist-mounted Basler acA1920-50gc camera.

\subsection{Insertion Task Domain Specification}
\label{sec:insertion_domain}

Here, we detail the domain specification for the insertion tasks, i.e., the 6-tuple $\langle S, \Phi, A, T, \Omega, O \rangle$ introduced in Sec.~\ref{subsec:domain}. We define the predicates to maximize the individual classifier performance of each sensor modality. While designing such predicates may not be straightforward for all manipulation tasks, we assume it is intuitive and unchallenging in a wide range of tasks, including insertion.

\subsubsection{States \texorpdfstring{$S$}{}}
See Fig.~\ref{fig:predicates}

\subsubsection{Predicates \texorpdfstring{$\Phi$}{}}
\begin{figure}
    \centering
    \includegraphics[width=0.95\columnwidth]{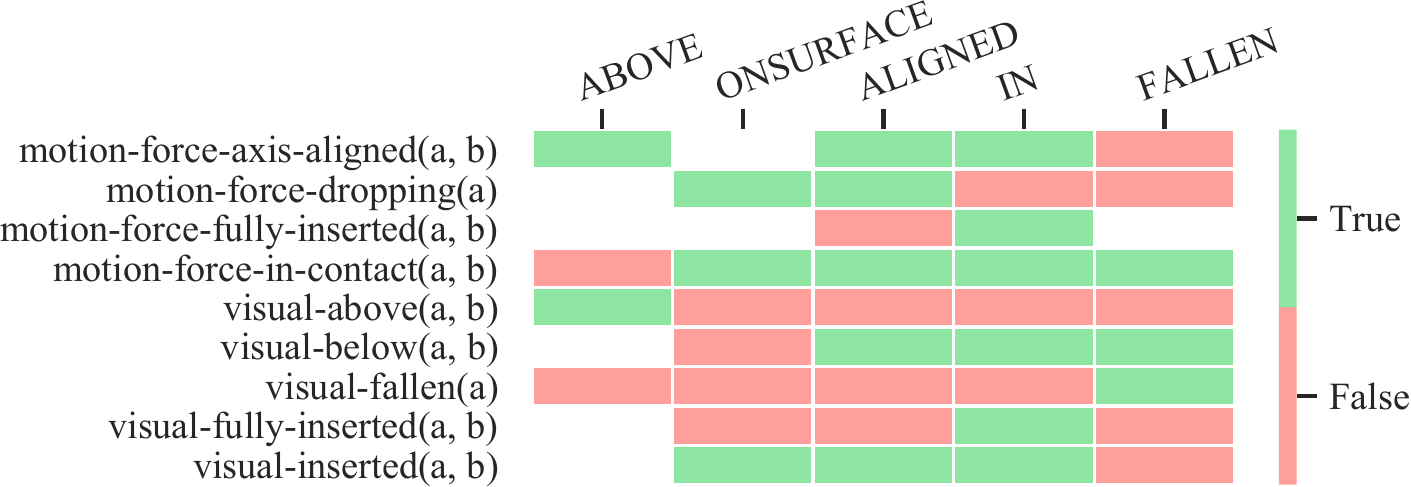}
    \caption{Symbolic states in insertion tasks and the corresponding predicate values. We define predicates to maximize the performance of each sensor modality. For example, it is easy for an image classifier to detect when a peg is positioned above or below the hole surface, but a motion and force-based classifier cannot easily tell without exact information about the surface height. One could choose to simply use a 1-to-1 mapping between states and predicates, but defining more nuanced predicates can simplify the classification problems and improve the overall performance.}
    \label{fig:predicates}
\end{figure}

See Fig.~\ref{fig:predicates}

\subsubsection{Actions \texorpdfstring{$A$}{}}
\begin{figure}
    \centering
    \includegraphics[width=\columnwidth]{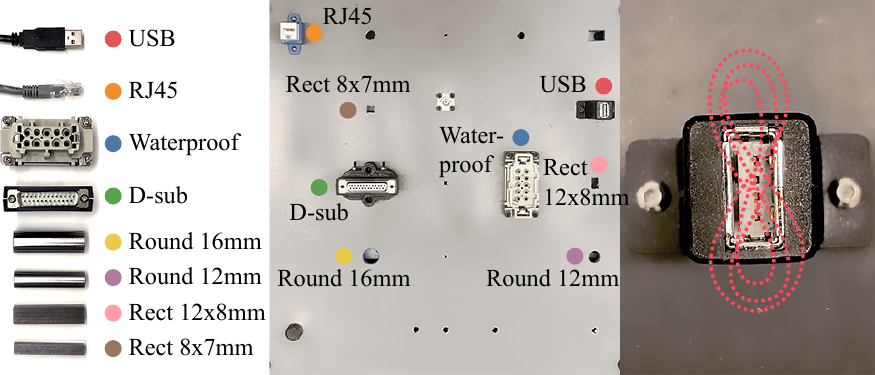}
    \caption{We evaluate our framework on 8 connector insertion tasks on the NIST Assembly Task Board I. We use a state machine with a set of predefined manipulation primitive skills to perform the task, such as Lissajous search (right) to find the connector hole.
    }
    \label{fig:nist}
\end{figure}

i) \textit{Prepare}: position the peg above the hole for insertion; ii) \textit{MakeContact}: move the peg down until touching the surface; iii) \textit{Search}: perform Lissajous search to find the hole (Fig.~\ref{fig:nist}); iv) \textit{Insert}: push the peg into the hole.

We implement these four actions using an impedance controller with different gain matrices ($K_P$ and $K_D$), reference pose ($x$), velocity ($\dot{x}$) and feed forward wrench ($F_{ff}$) profiles:
\begin{equation}
    \label{eq:controller}
    \tau = -J^T (K_P(x - x_d) + K_D(\dot{x} - \dot{x_d}) + F_{ff}),
\end{equation}
where $\tau$ is the control torque and $J$ is the Jacobian.

\subsubsection{Transitions \texorpdfstring{$T$}{}}
Computed from collected trajectories.

\subsubsection{Observations \texorpdfstring{$\Omega$}{}}
3D positions, 3D velocities, 3D forces, and $500 \times 500$ RGB images.

\subsubsection{Observation Models \texorpdfstring{$O$}{}}
Provided by the learned predicate classifiers and GMMs as detailed in Secs.~\ref{sec:bayesian} and \ref{sec:classifiers}.

\section{Experiments}
\label{sec:experiments}
We aim to investigate the following three questions in the experiments. First, we examine if introducing predicate classifiers boosts the symbolic state estimation performance. Second, we evaluate how well the state estimators generalize to unseen tasks of the same family. Third, we inspect whether the state estimators can improve the performance of a high-level manipulation policy by closing the loop on high-level states.

\subsection{Offline State Estimation}
\label{sec:offline}

\subsubsection{Experiment Setup}

For each insertion task, we execute $10$--$20$ trial runs with a predefined open-loop policy that uses manually tuned motion and force-based thresholds to complete each task. Specifically, the four actions (\textit{Prepare}, \textit{MakeContact}, \textit{Search}, and \textit{Insert}) in Sec.~\ref{sec:insertion_domain} are chained sequentially with duration limits. We then manually label the symbolic states for the collected observations and use these ground truth labels to evaluate the accuracy of four state estimation methods: 1) our proposed method using predicate classifiers (\textbf{Pred}), 2) a baseline trained to classify the symbolic state directly (\textbf{State}), 3) a differentiable filter adopting the categorical normalization strategy in \cite{zachares2020interpreting} (\textbf{Filter}), and 4) the predefined open-loop policy with manually tuned motion and force-based thresholds (\textbf{Manual}). We also perform an ablation study on the different sensor modalities to evaluate the impact of sensor fusion (\textbf{Pred-Image}, \textbf{Pred-MF}, \textbf{State-Image}, \textbf{State-MF}). We fuse sensor modalities in \textbf{State} by concatenating the learned features from each modality and feeding them to a multi-layer perceptron (MLP). We fuse sensor modalities in \textbf{Filter} by summing the conditional observation logits for each modality, analagous to Eq.~\ref{eq:fusion}.

For all methods except \textbf{Manual}, we perform $5$-fold cross validation using train-validation-test splits of $(0.6, 0.2, 0.2)$. To train the SimCLR image classifiers, we resample the dataset so it contains an equal number of samples from each symbolic state. Because sampled images are randomly perturbed during SimCLR training (Sec.~\ref{sec:simclr}), duplicating samples from rare symbolic states does not cause overfitting. Since differentiable filters are trained on fixed-length sequences of observations, we train \textbf{Filter} by sampling sequences of length 10 while ensuring that each symbolic state $s$ appears in at least $\frac{1}{|S|}$ of the training sequence set. All the images in one sequence undergo the same random SimCLR transformation. The models for all baselines are trained for $10$ epochs with batch sizes of $64$ using the same training hyperparameters.

\begin{figure}
    \centering
    \includegraphics[width=\columnwidth]{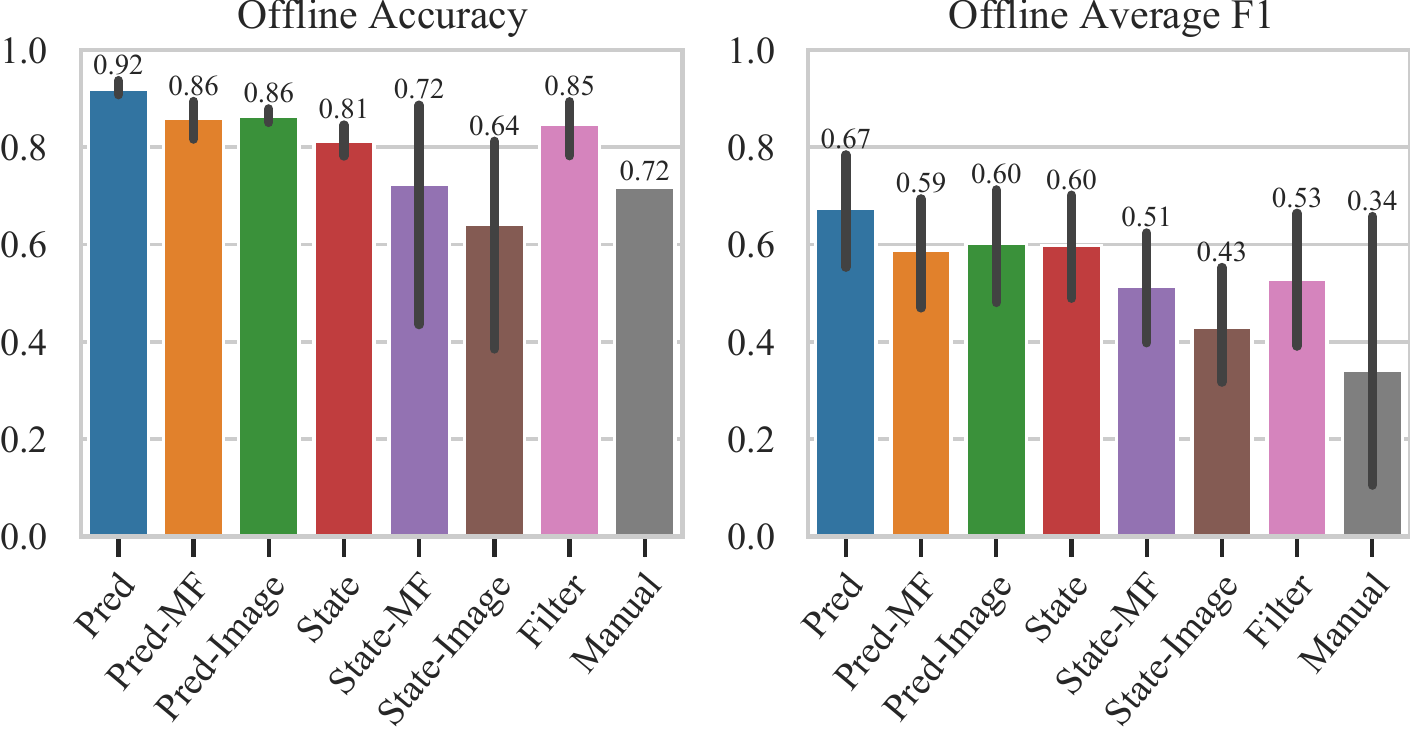}
    \caption{Offline evaluation of the state estimation methods. Left: the test accuracy computed via $5$-fold cross validation (except \textbf{Manual}, which requires no training). Right: the F1 score averaged across symbolic states; this metric particularly reflects the classification performance on the short duration states. Our method (\textbf{Pred}) performs the best on both metrics.}
    \label{fig:offline}
\end{figure}

\subsubsection{Offline Results}

As shown in Fig.~\ref{fig:offline}, our method (\textbf{Pred}) achieves the highest accuracy ($0.92$) and best F1 score averaged across all the states ($0.67$). The F1 score indicates how well the state estimator classifies short duration states, such as ALIGNED (when the connector drops slightly into the hole), which lasts less than $0.5$ seconds. These short states are crucial for insertion tasks---if the state estimator misclassifies the ALIGNED state, it may continue infinitely searching even if the connector is already aligned with the hole.

The fact that \textbf{Pred} outperforms \textbf{State} ($0.81$) in accuracy indicates that state estimation with the proposed binary predicate classifiers is more robust than with a direct multi-class symbolic state classifier. We also observe this performance gain when comparing the sensor ablations, (\textbf{Pred-MF} vs. \textbf{State-MF}) and (\textbf{Pred-Image} vs. \textbf{State-Image}).

\textbf{Filter} achieves a relatively high accuracy ($0.85$) but a low F1 score ($0.53$), indicating that it underperforms in classifying short duration states. A likely cause is the data imbalance of state occurrences. Although all short duration states are resampled and guaranteed to appear in at least $\frac{1}{|S|}$ of the training sequence set, the long duration states co-occur with those resampled sequences, leading to an overall domination by frames of long duration states. For \textbf{Pred} and \textbf{State}, on the other hand, the states are balanced at the frame granularity, rather than the sequence level, thus circumventing this issue.

The sensor modality ablations (\textbf{Pred-Image}, \textbf{Pred-MF}, \textbf{State-Image}, \textbf{State-MF}) all achieve lower accuracy and F1 scores than their corresponding combined versions (\textbf{Pred}, \textbf{Force}), indicating that fusing the sensor modalities increases reliability. The intuition is that motion/force and image signals often provide complementary information. For instance, force signals cannot distingiush between FALLEN and SEARCHING, while image-based SimCLR excels at this. Meanwhile, SimCLR struggles at distinguishing between ONSURFACE and ALIGNED, where the motion and force-based classifier shines. This is illustrated in Fig.~\ref{fig:teaser}, where the wrist camera images for ONSURFACE and ALIGNED are almost identical, but the velocity profiles are significantly different. Our proposed Bayesian estimator fuses multiple modalities by weighing their importance according to the individual classifier noise characteristics.

\subsection{Generalization to Unseen Tasks}
\label{sec:generalization}

\subsubsection{Experiment Setup}

In this experiment, we test the generalization ability of the state estimators to unseen tasks. To do so, we train each model on data collected from $7$ of the $8$ tasks and then test on the $8$th task. We perform this test for each task with $3$-fold cross validation. Since \textbf{Manual} requires no training, its results are the same as those in Sec.~\ref{sec:offline}.

\begin{figure}
    \centering
    \setlength{\abovecaptionskip}{0mm}
    \includegraphics[width=\columnwidth]{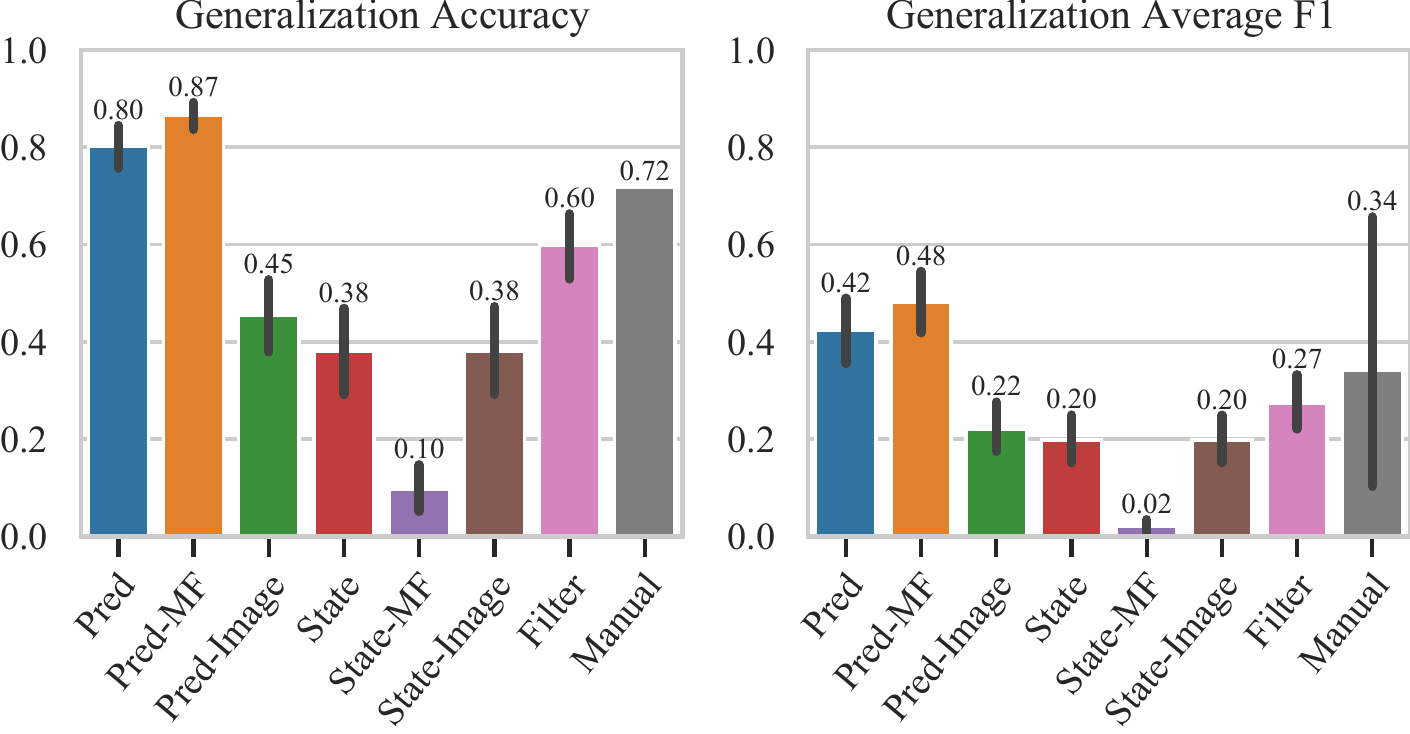}
    \caption{Generalization performance of the state estimation methods to unseen tasks. Both plots show the average result across all $8$ unseen tasks. \textbf{Pred} outperforms all other methods except \textbf{Pred-MF}, which demonstrates the benefit of using predicates for generalization. Meanwhile, image-based predicates are more challenging to generalize when trained on a limited dataset.}
    \label{fig:generalization}
\end{figure}

\subsubsection{Generalization Results}

As demonstrated in Fig.~\ref{fig:generalization}, \textbf{Pred} achieves higher accuracy and F1 scores than \textbf{State}, \textbf{Filter}, and \textbf{Manual}. The accuracy of \textbf{Pred} ($0.80$) decreases by only $0.12$ compared with Fig.~\ref{fig:offline}, where the method was trained on all the tasks. By contrast, the accuracy of \textbf{State} ($0.38$) drops by $0.43$. This drastic difference in accuracy indicates that state estimation with predicates offers stronger generalization to unseen tasks than direct state classification.

Among the predicate-based methods, \textbf{Pred-MF} achieves the best generalization performance. The reliability of \textbf{Pred-MF} on novel tasks suggests that it may be feasible to learn generic motion and force-based predicates across tasks in the same task family, e.g., connector-socket insertion. However, we expect weaker generalization for image-based predicates if not trained on a sufficiently large and diverse dataset, as indicated by \textbf{Pred-Image} in Fig.~\ref{fig:generalization}. Thus, we recommend using the trained predicate classifiers as a warm-start and further fine-tune on data collected from the novel task.

\subsection{Online Closed-Loop Policy Execution}
\label{sec:close-loop}

\subsubsection{Experiment Setup}

In this experiment, we integrate the state estimation methods with the connector insertion policy to close the high-level control loop; i.e., the state machine transitions are governed by the state estimates. We evaluate the four main methods (\textbf{Pred}, \textbf{State}, \textbf{Filter}, and \textbf{Manual}) on the three most error-prone insertion tasks (D-sub, USB, and RJ45)~\cite{lian2021benchmarking}. We assess the task success rate of each method on each connector over $20$ trial runs with varying hole positions within \SI{\pm 2}{\milli\meter} along all axes. For this experiment, we fine-tune the previously trained models with a subset of the dataset only containing the given insertion task using train-validation splits of $(0.75, 0.25)$, as suggested in Sec.~\ref{sec:generalization}.

The most common failure modes of the open-loop insertion policy include 1) the connector falling off the hole mount during SEARCHING, and 2) failing to detect the transition when the connector is ALIGNED with the hole. The state estimation methods can improve the performance of the predefined insertion policy by detecting these states and taking the appropriate high-level actions (e.g. resetting with a small perturbation or switching to insert). We enforce an execution time limit such that the state machine is terminated as a failure if it runs for more than one minute.

\begin{figure}
    \centering
    \setlength{\abovecaptionskip}{0mm}
    \includegraphics[width=\columnwidth]{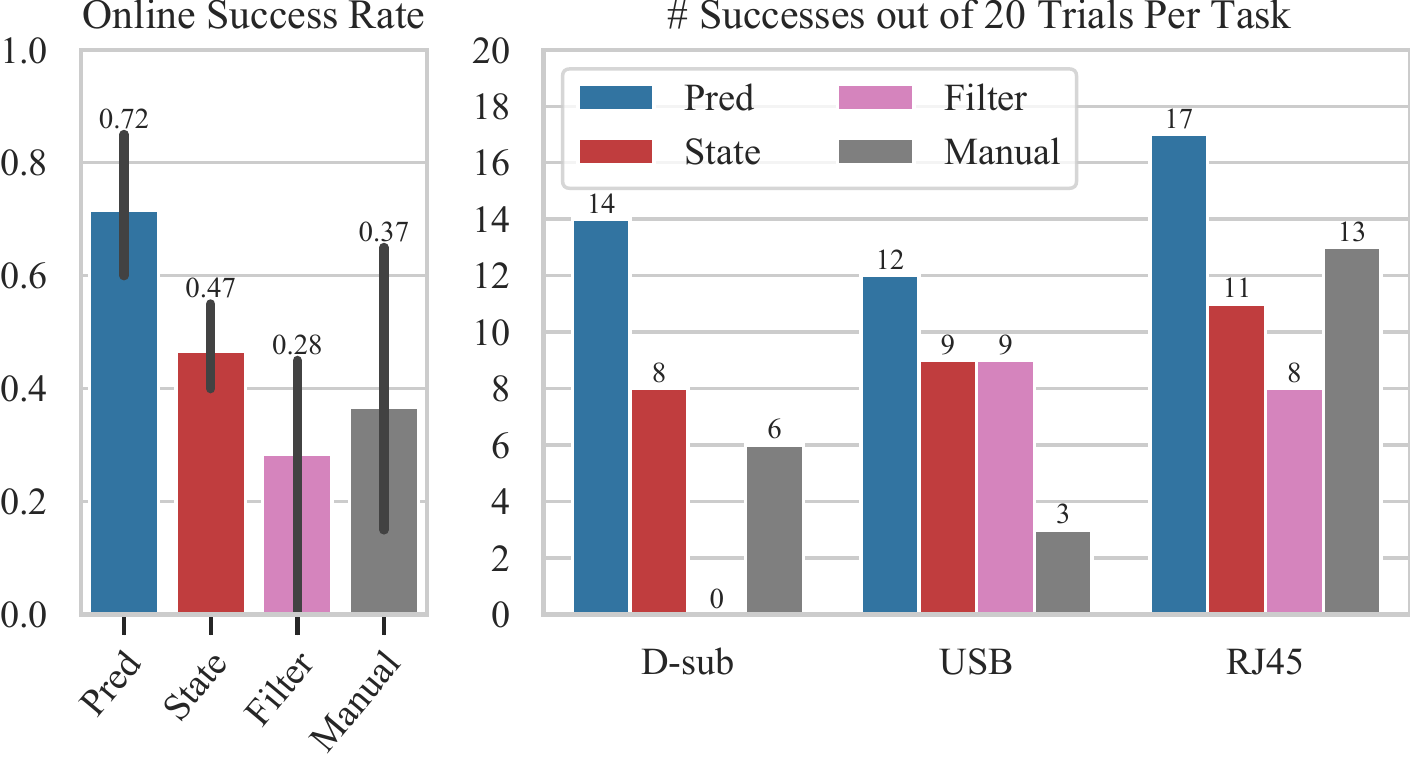}
    \caption{Online evaluation of the state estimation methods. Left: the average success rate across the $3$ tasks (D-sub, USB, RJ45). Right: the number of successes out of 20 trials for each task. Only \textbf{Pred} is able to outperform \textbf{Manual} on all tasks.}
    \label{fig:online}
\end{figure}

\subsubsection{Online Results}

The results are presented in Fig.~\ref{fig:online}, with demonstrations of policy executions provided in the supplementary video. \textbf{Pred} achieves the highest success rate in all $3$ tasks, $0.72$ on average. It is also the only method that consistently outperforms \textbf{Manual} for every task, demonstrating the promising performance gains by integrating our proposed estimator with an existing open-loop policy. Two most common failure modes of \textbf{Pred} include 1) when the connector keeps falling off the hole mount after new resets, reaching the execution time limit, and 2) when SEARCHING takes too long to find the hole, leading to time out. There are few failures due to misclassification by \textbf{Pred}, and we leave optimizing low-level execution policies to future work.

\textbf{Filter} fails in all trials of the D-sub task because it cannot detect when the connector is ALIGNED with the hole and keeps searching even if the connector is fully inserted. This again confirms our earlier hypothesis on its inability to handle short duration states such as ALIGNED.

\section{Conclusion}
\label{sec:conclusion}

This paper presents a Bayesian framework that uses binary predicate classifiers to estimate high-level symbolic states from high-dimensional sensor modalities in a data-efficient manner. This method outperforms other baselines such as direct symbolic state classification, differentiable filtering, and manually defined thresholds. We demonstrate with a real-world connector insertion task that symbolic state estimation can improve the performance of high-level policies by detecting and recovering from failures and deciding state transitions. Although we apply our method to state machines, symbolic state estimators could easily be used with other high-level execution models such as behavior trees and task planners.


\bibliographystyle{IEEEtranN}
\bibliography{references}


\end{document}